\documentclass{article}

\usepackage[final,nonatbib]{flfm_neurips_2023}

\usepackage{multirow}
\usepackage{algorithm2e}
\RestyleAlgo{ruled}

\usepackage{caption}
\usepackage{subcaption}
\usepackage{booktabs}
\usepackage{arydshln}

\usepackage{amsmath,amsfonts,bm}

\def\eqref#1{equation~\ref{#1}}

\def\1{\bm{1}}

\def\rvg{{\mathbf{g}}}
\def\rvh{{\mathbf{h}}}

\def\rvw{{\mathbf{w}}}
\def\rvx{{\mathbf{x}}}
\def\rvy{{\mathbf{y}}}
\def\rvz{{\mathbf{z}}}

\def\vmu{{\bm{\mu}}}
\def\vtheta{{\bm{\theta}}}

\def\vepsilon{{\bm{\epsilon}}}

\def\mI{{\bm{I}}}

\DeclareMathAlphabet{\mathsfit}{\encodingdefault}{\sfdefault}{m}{sl}
\SetMathAlphabet{\mathsfit}{bold}{\encodingdefault}{\sfdefault}{bx}{n}

\def\gN{{\mathcal{N}}}

\newcommand{\E}{\mathbb{E}}

\DeclareMathOperator*{\argmax}{arg\,max}
\DeclareMathOperator*{\argmin}{arg\,min}

\usepackage[utf8]{inputenc} %
\usepackage[T1]{fontenc}    %
\usepackage{hyperref}       %
\usepackage{url}            %
\usepackage{booktabs}       %
\usepackage{amsfonts}       %
\usepackage{nicefrac}       %
\usepackage{microtype}      %
\usepackage{xcolor}         %
\usepackage{graphicx}
\usepackage{mathtools}

\newcommand{\methodName}{DiffULA}

\title{Exploring User-level Gradient Inversion \\ with a Diffusion Prior}

\author{%
  Zhuohang Li\\
  Vanderbilt University\\
  \texttt{zhuohang.li@vanderbilt.edu} \\
  \And
  Andrew Lowy\\
  University of Wisconsin-Madison \\
  \texttt{alowy@wisc.edu} \\
  \And
  Jing Liu\\
  Mitsubishi Electric Research Laboratories \\
  \texttt{jiliu@merl.com} \\
  \And
  Toshiaki Koike-Akino\\
  Mitsubishi Electric Research Laboratories \\
  \texttt{koike@merl.com} \\
  \AND
  Bradley Malin\\
  Vanderbilt University Medical Center \\
  \texttt{b.malin@vumc.org} \\
  \And
  Kieran Parsons\\
  Mitsubishi Electric Research Laboratories \\
  \texttt{parsons@merl.com} \\
  \And
  Ye Wang\\
  Mitsubishi Electric Research Laboratories \\
  \texttt{yewang@merl.com} \\
}

\begin{document}

\maketitle

\begin{abstract}
We explore user-level gradient inversion as a new attack surface in distributed learning. We first investigate existing attacks on their ability to make inferences about private information beyond training data reconstruction. Motivated by the low reconstruction quality of existing methods, we propose a novel gradient inversion attack that applies a denoising diffusion model as a strong image prior in order to enhance recovery in the large batch setting. Unlike traditional attacks, which aim to reconstruct individual samples and suffer at large batch and image sizes, our approach instead aims to recover a representative image that captures the sensitive shared semantic information corresponding to the underlying user. Our experiments with face images demonstrate the ability of our methods to recover realistic facial images along with private user attributes.

\end{abstract}

\section{Introduction}

Computing and storing \textit{gradients} is essential for training most types of modern deep-learning models, ranging from compact neural networks~\cite{howard2017mobilenets} designed for edge applications to massive foundational models~\cite{bommasani2021opportunities} that are fine-tuned using gradient descent.
In circumstances such as distributed training and federated learning~\cite{mcmahan2017communication}, gradients are directly exchanged in place of raw training data to facilitate joint learning from large-scale distributed data.

Plainly sharing gradient information was long presumed to be safe.
However, \textit{gradient inversion}, a new type of privacy threat was recently demonstrated that calls this assumption into question. Zhu \textit{et al.}~\cite{zhu2019deep} demonstrated that an adversary can recover the underlying private training data via an iterative gradient-matching optimization procedure.
While plausible, this optimization problem becomes ill-posed and difficult to solve when the data is high-dimensional~\cite{geiping2020inverting} and from a large batch size~\cite{yin2021see}, or if local defenses are applied to obfuscate the gradients~\cite{li2022auditing}.
As a result, existing solutions based on total variation~\cite{geiping2020inverting} or generative adversarial network (GAN) based image priors~\cite{jeon2021gradient, li2022auditing} are prone to degradation in their reconstructed image quality with decreasing amounts of details as batch size and image resolution increase.
Further, GAN-based approaches are also known to suffer from mode collapse~\cite{bayat2023study}, which can result in limited sample diversity and degrade overall attack performance.

In this work, we explore \textit{user-level} gradient inversion as a new attack surface in distributed learning, beyond the risks of training data recovery.
This is motivated by collaborative learning settings, where data across users may be neither independent nor identically distributed, and particularly when the data from each user may consistently belong to a distinct individual.
Thus, while recovering individual data samples may be difficult, the potential inference of sensitive user-level characteristics may still pose a risk.

We empirically investigate the user-level privacy risks of existing gradient inversion methods, revealing their abilities to recover some level of private information albeit with low image quality.
To improve reconstructed image quality and enable attacks against large batch sizes,
we also propose a new type of user-level gradient inversion attack, utilizing a pre-trained diffusion model prior to assist the reconstruction of meaningful images.
Specifically, in contrast to existing sample-level attacks that attempt to fully reconstruct every image in the target batch, we instead aim to synthesize a single representative image that captures the overall semantics of the private image batch, which makes the search space dimension invariant to the target batch size and thereby significantly reduces computational overheads and improves convergence stability.

Our method is motivated by the recent success in the research domain of using diffusion priors to solve inverse problems~\cite{song2021solving, kawar2022denoising, chung2022improving, chung2023diffusion}.
However, most existing research focuses on typical image inverse problems, such as denoising and inpainting, which have well-defined linear forward operators. By contrast, the forward process in our problem involves the calculation of model gradients, which involves a complex series of non-linear operations.
Moreover, the measurements in gradient inversion are the observed gradients, which contrasts with traditional inverse problems, where both the original and measured signals are in the image space, rendering many existing methods inapplicable.
To address these challenges, we apply a plug-and-play diffusion prior technique~\cite{graikos2022diffusion} that formulates the inference of the posterior distribution as an optimization problem
and propose a dynamic optimization scheme that focuses on recovering the high-level semantics at early stages and then fine-tune image details at the later stage.
To the best of our knowledge, this is the first study to investigate the application of a diffusion model prior for improving gradient inversion.

\begin{figure}[t]
    \centering
    \includegraphics[width=0.8\textwidth]{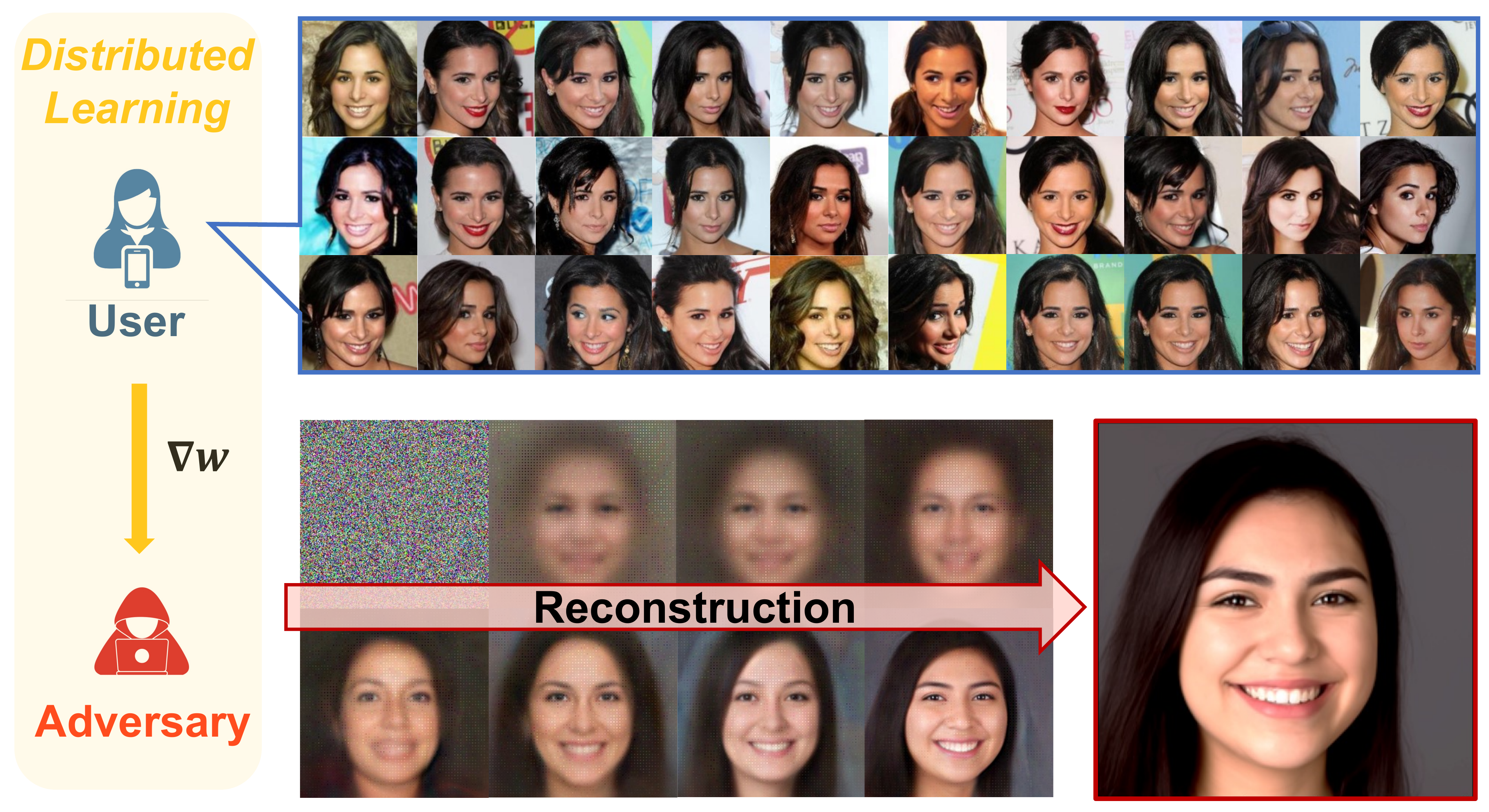}

    \caption{Illustration of \textit{user-level} gradient inversion with diffusion prior. \textbf{Top}: user's private batch of $30$ images. The measured pairwise facial similarity in the private batch ranges from 0.5217 to 0.9579. \textbf{Bottom}: reconstructed image from gradients using the proposed method. The average facial similarity to the original batch is 0.5565.}
    \label{fig:enter-label}

\end{figure}

Distinct from the prior literature, our evaluation extends beyond the conventional performance metrics for gradient inversion which measure only image-level similarity to inspect the recovery of certain private attributes that are more semantically meaningful and interpretable.
Through experiments on the CelebA facial image dataset~\cite{liu2015faceattributes}, we show that the proposed user-level gradient inversion method can effectively recover high-level semantics of large-batches of images that carry important private information about the person, including gender, race, age, and facial identity, without relying on strong adversarial assumptions such as the BatchNorm statistics~\cite{huang2021evaluating}.

\section{User-level Gradient Inversion with Diffusion Prior}

\subsection{Problem Formulation}

We consider the typical supervised distributed learning setting with a set of clients $C$, where the goal is to find the optimal model parameters ${\rvw}$ of a neural network $f_{\rvw}$ such that the empirical risk is minimized, i.e., $\min_{\rvw} \sum_{c \in C} \sum_{(\rvx_i, \rvy_i) \in \mathcal{D}_c} \mathcal{L}(f_{\rvw}(\rvx_{(i)}, \rvy_{(i)})$, where $\rvx_{(i)}, \rvy_{(i)}$ denotes the $i$-th image and corresponding label, respectively, $\mathcal{D}_c$ represents the client's local training dataset, and $\mathcal{L}$ is the loss function (e.g., cross-entropy). The most common practice is to optimize ${\rvw}$ through stochastic gradient descent. Specifically, at each training step, the client node samples a batch of $B$ images $\{(\rvx_{(i)}, \rvy_{(i)})\}_{i=1}^B$ to compute the averaged gradients ${\nabla \rvw} = \frac{1}{B}\sum_{i=1}^B \nabla_{\rvw} \mathcal{L}(f_{\rvw}(\rvx_{(i)}), \rvy_{(i)})$, which are then transferred over the communication channel to the server (in the parameter server setting) or its peers (in the peer-to-peer setting). In the gradient inversion attack, the adversary
observes the user's gradients ${\nabla \rvw}$ and global model weights $\rvw$ and tries to recover the original images and labels.
When the adversary's access is limited to aggregated gradients (e.g., under secure aggregation), they may first initiate a disaggregation attack~\cite{lam2021gradient} to obtain gradients at the user level.
Following prior work~\cite{zhao2020idlg,geiping2020inverting,yin2021see,jeon2021gradient,li2022auditing}, we assume the labels can be analytically recovered from the gradient of the last layer. Therefore, the images can be reconstructed by solving the following optimization problem:
\begin{equation}\label{eq:grad_match}
\hat{\rvx}_{(1)}^*, ..., \hat{\rvx}_{(B)}^* = \argmin_{\hat{\rvx}_{(1)}, ..., \hat{\rvx}_{(B)}} \mathbf{d}(F(\hat{\rvx}_{(1)}, ..., \hat{\rvx}_{(B)}), {\nabla \rvw}), 
\end{equation}
where $\mathbf{d}$ is a distance metric (e.g., Euclidean squared distance~\cite{zhu2019deep} or cosine distance~\cite{geiping2020inverting}),  $\hat{\rvx}_{(1)}, ..., \hat{\rvx}_{(B)}$ is a batch of synthetic data, and $F(\hat{\rvx}_{(1)}, ..., \hat{\rvx}_{(B)}) = \frac{1}{B}\sum_{i=1}^B \nabla_{\rvw} \mathcal{L}(f_{\rvw}(\hat{\rvx}_{(i)}), \rvy_{(i)})$ is the corresponding gradient. This technique is referred to as \textit{gradient matching}~\cite{zhu2019deep} in the literature and it has been empirically shown that as the distance in the gradient space reduces, the synthetic images will gradually recover the original images.

\subsection{User-Level Gradient Inversion}

\subsubsection{Challenge of Sample-level Reconstruction}

The gradient inversion attack in its canonical form (Eq.~\ref{eq:grad_match}) suffers poor scalability in multiple aspects. First, due to the ill-posedness of the inverse problem, the returned synthetic image through naively minimizing the gradient matching loss may not be a natural image. These effects become more significant on deeper networks and higher-resolution images. One mitigation is to instead solve:
\begin{equation}\label{eq:reg}
\hat{\rvx}_{(1)}^*, ..., \hat{\rvx}_{(B)}^* = \argmin_{\hat{\rvx}_{(1)}, ..., \hat{\rvx}_{(B)}} \mathbf{d}(F(\hat{\rvx}_{(1)}, ..., \hat{\rvx}_{(B)}), {\nabla \rvw}) + \mathcal{R}_\text{prior}(\hat{\rvx}_{(1)}, ..., \hat{\rvx}_{(B)}), 
\end{equation}
where $\mathcal{R}_\text{prior}$ is an additional prior term (e.g., total variation~\cite{geiping2020inverting}) introduced to regularize the synthetic image. A more recent line of work considers using the generator $G$ from pre-trained GANs~\cite{jeon2021gradient,li2022auditing} as a regularization and solve for low-dimensional latent vectors $\rvz_1, \rvz_2, ..., \rvz_B$ that minimizes the gradient matching loss, i.e.,
\begin{equation}
\hat{\rvz}_{(1)}^*, ..., \hat{\rvz}_{(B)}^* = \argmin_{\hat{\rvz}_{(1)}, ..., \hat{\rvz}_{(B)}} \mathbf{d}(F(G(\hat{\rvz}_{(1)}), ..., G(\hat{\rvz}_{(B)})), {\nabla \rvw}). 
\end{equation}
Although this form enables the gradient inversion to generalize to larger images, it still does not address the other challenge that the search space scales linearly with the batch size $B$ and the attack performance quickly degrades as $B$ increases. As such, most existing gradient inversion attacks are confined to small-sized images with relatively small batch sizes (e.g.,~\cite{jeon2021gradient} considered a batch of $4$ images rescaled to the size of $64 \times 64 \text{px}$). The only exception is the work by Yin \textit{et al.}~\cite{yin2021see}, which is based on the strong assumption that the clients share additional BatchNorm statistics beyond regular gradients~\cite{huang2021evaluating}.
Another common assumption by prior work is that images in the batch are uncorrelated~\cite{kariyappa2023cocktail} or from different classes~\cite{yin2021see}, which may not hold when learning across non-IID data.
Extending gradient inversion attacks to practical batch sizes without such strong assumptions remains a challenging research problem~\cite{huang2021evaluating}.

\subsubsection{From Sample-level to User-level Inversion}

In the sample-level gradient inversion attacks as described above, the adversary observes the gradients ${\nabla \rvw}$ computed from a batch of $B$ images, and the goal is to reconstruct every sample in the original batch, which is equivalent to estimating images to maximize $p(\rvx_1, \rvx_2, ..., \rvx_B|{\nabla \rvw})$~\cite{balunovic2022bayesian}.
However, in many practical distributed learning scenarios, such as cross-device federated learning on mobile devices~\cite{hard2018federated,kairouz2021advances}, data may often have strong local correlation (and may not be identical between users), i.e., local data samples are from the same user and thus have similar semantics.
In these settings, the ultimate goal of a practical attack may not be image reconstruction, but rather recovery of private information about the user, such as demographics and identity.
To account for this semantic consistency across local batches, we extend the Bayesian model of the data to introduce a latent encoding $\rvh_c$ %
that captures the user demographics that the adversary wishes to recover (e.g., gender, age, race, etc.). %
Joint recovery of the images and latent encoding, given a fixed $\nabla \rvw$, is guided by the probabilistic model
\begin{align}
p(\rvx_{(1)}, ..., \rvx_{(B)}, \rvh_c | {\nabla \rvw}) & \propto p(\rvx_{(1)}, ..., \rvx_{(B)}, \rvh_c, {\nabla \rvw}) \\
&= p({\nabla \rvw} | \rvx_{(1)}, ..., \rvx_{(B)}) p(\rvh_c | \rvx_{(1)}, ..., \rvx_{(B)}) p(\rvx_{(1)}, ..., \rvx_{(B)}),
\end{align}
where the equality is due to $\rvh_c \to (\rvx_{(1)}, ..., \rvx_{(B)}) \to {\nabla \rvw}$ forming a Markov chain, and which suggests the reconstruction optimization
\begin{align}\label{eq:user_level_objective}
\argmax_{\rvx_{(1)}, ..., \rvx_{(B)}, \rvh_c} \underbrace{\log p({\nabla \rvw}|\rvx_{(1)}, ..., \rvx_{(B)})}_\text{grad matching} + \underbrace{\log p(\rvh_c | \rvx_{(1)}, ..., \rvx_{(B)})}_\text{consistency} + \underbrace{\log p(\rvx_{(1)}, ..., \rvx_{(B)})}_\text{image prior}.
\end{align}
Hence, the adversary should consider three terms, where the third term is the image prior, which will be discussed in the next section.
The first term implies the recovered images should produce the observed gradient, which is achievable through gradient matching.
The second term implies that the images should be consistent with the latent encoding, i.e., sharing semantics due to being from the same user.

Exploiting this semantic similarity, while obtaining computational advantages, the adversary can instead aim to recover a single representative image $\hat{\rvx}$ that captures the semantics of the original batch $\{\rvx_{(i)} \}_{i=1}^B$ by solving
\begin{equation}\label{eq:user_attack}
\hat{\rvx}^* = \argmin _{\hat{\rvx}} - \log p({\nabla \rvw}|\mathcal{A}(\hat{\rvx})) - \log p(\hat{\rvx}),
\end{equation}
where $\mathcal{A}(\hat{\rvx})=\{\mathcal{T}(\hat{\rvx}) \}_{i=1}^B$ is a batch of $B$ augmented images produced from $\hat{\rvx}$ with random semantics-preserving image transformation(s) $\mathcal{T}$ (i.e., $p(\rvh_c|\mathcal{T}(\hat{\rvx})) \equiv p(\rvh_c|\hat{\rvx})$), and use $\argmax_{\rvh_c} p(\rvh_c|\hat{\rvx})$ as the predicted value for $\rvh_c$.
This approach has the benefits of simplifying the reconstruction of a potentially large batch to a single image, which substantially reduces the computational cost and memory consumption, accelerates the optimization, and thereby enables the attack to be extended to larger batch sizes.
Despite this, in practice optimizing the first term in Eq.~\ref{eq:user_attack} drives the synthetic image $\hat{\rvx}$ towards the direction where its produced gradient approximates the observed gradient computed from a batch of images, which is likely to result in $\hat{\rvx}$ converging to an unrealistic noisy image. To obtain a natural image, we need to employ a strong prior to regularize $\hat{\rvx}$ during the optimization process, which we introduce next.

\subsection{Diffusion Prior for Gradient Inversion}

\subsubsection{Denoising Diffusion Probabilistic Model as Prior}

The denoising diffusion probabilistic model (DDPM)~\cite{ho2020denoising} is a generative model based on a Markov chain capturing a forward diffusion process that progressively adds Gaussian noise, given by
\begin{equation}
q(\rvx_{1:T}|\rvx_0)=\prod_{t=1}^T q(\rvx_t|\rvx_{t-1}), \quad q(\rvx_t|\rvx_{t-1}) = \gN(\rvx_{t};\sqrt{1-\beta_t}\rvx_{t-1}, \beta_t\mI),
\end{equation}
where $\rvx_0 := \rvx \sim q$ represents a data sample from the distribution being modelled, and $\beta_t \in (0, 1)$ are fixed noise schedule hyper-parameters.
This forward process can be summarized as
\begin{equation}\label{eq:sample_xt}
    q(\rvx_t|\rvx_0) = \gN(\rvx_t; \sqrt{\bar{\alpha}_t}\rvx_0, (1-\bar{\alpha}_t)\mI),
\end{equation}
where $\bar{\alpha}_t := \prod_{s=1}^t \alpha_s$, and $\alpha_t := 1-\beta_t$.
The forward process contains no trainable parameters, and DDPMs ultimately model the data distribution through a parameterized reverse process, given by
\begin{equation}
p_\vtheta(\rvx_{0:T}) = p(\rvx_T) \prod_{t=1}^T p_\vtheta(\rvx_{t-1}|\rvx_t), \quad p_\vtheta(\rvx_{t-1}|\rvx_t) = \gN(\rvx_{t-1};\vmu_\vtheta(\rvx_t, t), \sigma_t^2 \mI)
\end{equation}
where $p(\rvx_T) = \gN(\bm{0}, \mI) \approx q(\rvx_T)$, $\sigma_t$ is specified as a function of $\beta_t$, and %
\begin{equation}
\vmu_\vtheta(\rvx_t, t) := \frac{1}{\sqrt{\alpha_t}} \left( \rvx_t - \frac{\beta_t}{\sqrt{1 - \bar{\alpha}_t}} \vepsilon_\vtheta(\rvx_t, t) \right),
\end{equation}
where $\vepsilon_\vtheta$ is a neural network parameterized by $\vtheta$ that is essentially trained to denoise, i.e., by recovering $\big( (\rvx_t - \sqrt{\bar{\alpha}_t} \rvx_0) / \sqrt{1 - \bar{\alpha}_t} \big)$, given samples $\rvx_t$ drawn from the forward process in Eq.~\ref{eq:sample_xt}.

The training of a DDPM (see~\cite{ho2020denoising} for further details) involves minimizing a variational upper bound of the negative log likelihood of the model,
\begin{equation}
\E_q[- \log p_\vtheta(\rvx_0)] \leq \E_q \left[- \log \frac{p_\vtheta(\rvx_{0:T})}{q(\rvx_{1:T}|\rvx_0)} \right] =: L.
\end{equation}
With further derivation and discarding terms that are constant with respect to the trainable parameters (see~\cite{ho2020denoising}), this likelihood bound can be expressed as
\begin{equation} \label{eq:mod_bound}
\sum_{t} w_t \E_{\rvx_0 \sim q, \vepsilon \sim \gN(\bm{0}, \mI)}\Big[ \big\|\vepsilon - \vepsilon_\vtheta \big(\sqrt{\bar{\alpha}_t}\rvx_0 + \sqrt{1-\bar{\alpha}_t}\vepsilon, t \big) \big\|^2 \Big],
\end{equation}
where the weights $w_t$ are a function of the noise schedule $\beta_t$.

Similar to the procedure of~\cite{graikos2022diffusion}, we use Eq.~\ref{eq:mod_bound} as an effective proxy for the model likelihood, by evaluating it with respect to $q(\rvx_0) = \delta(\rvx_0-\hat{\rvx}_0)$, i.e., the delta distribution centered on the estimate variable $\hat{\rvx}_0$.
Combining this into Eq.~\ref{eq:user_attack}, we obtain an overall inversion strategy employing a pre-trained DDPM model as a prior, given by
\begin{equation}\label{eq:final}
\hat{\rvx}^* = \argmin _{\hat{\rvx}} - \log p({\nabla \rvw}|\mathcal{A}(\hat{\rvx})) - \sum_{t} w_t \E_{\vepsilon \sim \gN(\bm{0}, \mI)}\Big[ \big\|\vepsilon - \vepsilon_\vtheta \big(\sqrt{\bar{\alpha}_t}\hat{\rvx} + \sqrt{1-\bar{\alpha}_t}\vepsilon, t \big) \big\|^2 \Big].
\end{equation}

\subsubsection{Optimization}

In our implementation, we employ cosine distance~\cite{geiping2020inverting} as a proxy for the gradient matching term in Eq.~\ref{eq:final}, which allows the overall optimization to be solved with standard gradient descent. 
However, due to the stochasticity of the reverse diffusion process, in practice the optimization becomes stochastic and may produce different estimates from different runs that could potentially deviate from the observed gradient.
To ensure the production of reliable results that satisfy both the diffusion prior and the gradient constraint, ideally, the early stage of the optimization should coarsely explore the search space, while the later stage of the optimization should carefully fine-tune the estimated $\hat{\rvx}_0$ to steadily converge to a local maximum~\cite{graikos2022diffusion}.

To this end, we make the following refinements to the optimization process. First, we observe that to achieve the best recovery, the gradient matching loss should be adjusted according to the phase of the optimization: in the early stage, we should first focus on recovering the correct high-level semantics such as gender and race; then the later stage we can extend to reconstruct detailed facial attributes.
Therefore, instead of assigning equal weights to all gradients, we apply a sliding asymmetric Hamming window function to dynamically adjust the weight of the gradient of each layer. Second, instead of simultaneously optimizing for all $t$, we optimize according to a schedule that gradually anneals from $T$ to a small value $t_\text{min}$. Third, similar to~\cite{ho2020denoising}, we use constant weights $w_t$ to simplify the optimization objective. To balance the two terms, we instead clip the gradient norm of the gradient matching loss dynamically according to the gradient norm of the diffusion prior loss term.
The full algorithm and a detailed description of the procedure are included in the Appendix.

\section{Experiments}

\paragraph{Setup.}
We consider the binary classification task of attractiveness on the CelebA dataset~\cite{liu2015faceattributes}. The images are resized to $64\times64$ px and gradients are computed on a randomly initialized ResNet-18 with a default batch size of $30$. To simulate cross-device FL settings, images are grouped according to user ID. The experiments use the first $50$ users with $\geq 30$ images as the evaluation set.

\begin{table}[t]
\caption{Numerical comparison of the proposed method and baselines on the CelebA dataset.}
\begin{subtable}[c]{0.39\linewidth}
\centering

\caption{Image quality}

\resizebox{\linewidth}{!}{
\begin{tabular}{cccc}
\toprule
& \textbf{MSE$\downarrow$} & \textbf{PSNR$\uparrow$} & \textbf{LPIPS$\downarrow$} \\ \toprule
\textit{\textbf{Intra-user}}                                                           & 0.4981                   & 3.7879                  & 0.4392                     \\ \midrule
\textit{\textbf{Inverting}}                                                           & 0.5163                   & 3.2927                  & 0.7324                     \\ \hdashline
\textit{\textbf{\begin{tabular}[c]{@{}c@{}}w/ disambiguation\end{tabular}}} & 0.3192                   & 5.2633                  & 0.7165                     \\ \midrule
\textit{\textbf{GIAS}}                                                                & 0.5024                   & 3.5402                  & 0.6173                     \\ \hdashline
\textit{\textbf{\begin{tabular}[c]{@{}c@{}}w/ disambiguation\end{tabular}}}      & 0.2820                   & 5.8809                  & 0.5887                     \\ \midrule
\textit{\textbf{DiffULA}}                                                            & 0.4210                   & 4.6789                  & 0.5039                     \\ \bottomrule
\end{tabular}
}
\label{tab:res_quality}
\end{subtable}
\hfill
\begin{subtable}[c]{0.6\linewidth}
\centering

\caption{Semantics}

\resizebox{\linewidth}{!}{
\begin{tabular}{cccccc}
\toprule
& \textbf{\begin{tabular}[c]{@{}c@{}}Face\\ Detection Rate$\uparrow$\end{tabular}} & \textbf{\begin{tabular}[c]{@{}c@{}}Facial\\ Similarity$\uparrow$\end{tabular}} & \textbf{\begin{tabular}[c]{@{}c@{}}Gender\\ Accuracy$\uparrow$\end{tabular}} & \textbf{\begin{tabular}[c]{@{}c@{}}Race\\ Accuracy$\uparrow$\end{tabular}} &  \textbf{\begin{tabular}[c]{@{}c@{}}Age\\ Error$\downarrow$\end{tabular}} \\ \toprule
\textit{\textbf{Original Batch}}                                                      & 0.95                                                                            & 0.7112                                                                         & 0.94                                                                         & 0.89                                                                                                                                                 & 2.37                                                                     \\ \midrule
\textit{\textbf{Inverting}}                                                      & 0.03                                                                            & 0.3939                                                                         & 0.16                                                                         & 0.12                                                                                                                                                & 2.54                                                                     \\ \hdashline
\textit{\textbf{\begin{tabular}[c]{@{}c@{}}w/ ensemble\end{tabular}}} & (0.22)                                                                             & 0.1275                                                                         & 0.44                                                                         & 0.48                                                                                                                                                 & 4.91                                                                    \\ \midrule
\textit{\textbf{GIAS}}                                                           & 0.17                                                                            & 0.5986                                                                         & 0.18                                                                         & 0.84                                                                                                                                                 & 2.64                                                                     \\ \hdashline
\textit{\textbf{\begin{tabular}[c]{@{}c@{}}w/ ensemble\end{tabular}}}      & (0.96)                                                                             & 0.6129                                                                         & 0.52                                                                         & 0.80                                                                                                                                                 & 2.90                                                                     \\ \midrule
\textit{\textbf{\methodName}}                                                       & 1.00                                                                            & 0.5740                                                                         & 0.71                                                                         & 0.29                                                                                                                                                 & 3.61                                                                     \\ \bottomrule
\end{tabular}
}
\label{tab:res_seman}
\end{subtable}

\end{table}

\paragraph{Baselines.} We compare user-level attack with diffusion prior (\textit{\methodName}) with two baselines, namely, \textit{inverting gradients}~\cite{geiping2020inverting} based on total variation image prior and \textit{GIAS}~\cite{jeon2021gradient} based on GAN prior. We set hyperparameters by following original works~\cite{geiping2020inverting,jeon2021gradient}. For GIAS, we use the state-of-art Diffusion-StyleGAN~\cite{wang2023diffusion} pretrained on CelebA for generating $64\times64$ px images. We note that this is an unrealistically strong threat model, as the private evaluation images were included in the GAN training, however this can be viewed as a conservative upper bound on this baseline performance. For the diffusion prior, we use the DDPM weights from prior work~\cite{baranchuk2022label}, which are pre-trained on the FFHQ dataset~\cite{karras2019style} for generating $256\times256$ px images.

\paragraph{Evaluation Metrics.}
Following prior works~\cite{geiping2020inverting,yin2021see,jeon2021gradient,li2022auditing}, we use image space Mean-Squared Error (MSE), Peak Signal-to-Noise Ratio (PSNR), and Learned Perceptual Image Patch Similarity (LPIPS)~\cite{zhang2018unreasonable} as image quality metrics. Additionally, we introduce the following metrics for measuring the semantic similarity between the original and reconstructed images: (1) \textit{Face Detection Rate}: ratio of reconstructions with at least one image that can be successfully detected by an MTCNN~\cite{zhang2016joint} face detector; (2) \textit{Gender Accuracy}: ratio of reconstructions with correct gender; (3) \textit{Race Accuracy}: ratio of reconstructions with correct race; (4) \textit{Age Error}: average error between the age of original images and the reconstructed images;
(5) \textit{Facial Similarity}: the similarity between the original and the reconstructed images measured in the embedding space. We use pre-trained neural networks provided in the DeepFace~\cite{serengil2020lightface,serengil2021lightface} package for evaluating (1)-(5).

\begin{figure}[t]
    \centering
    \includegraphics[width=0.86\textwidth]{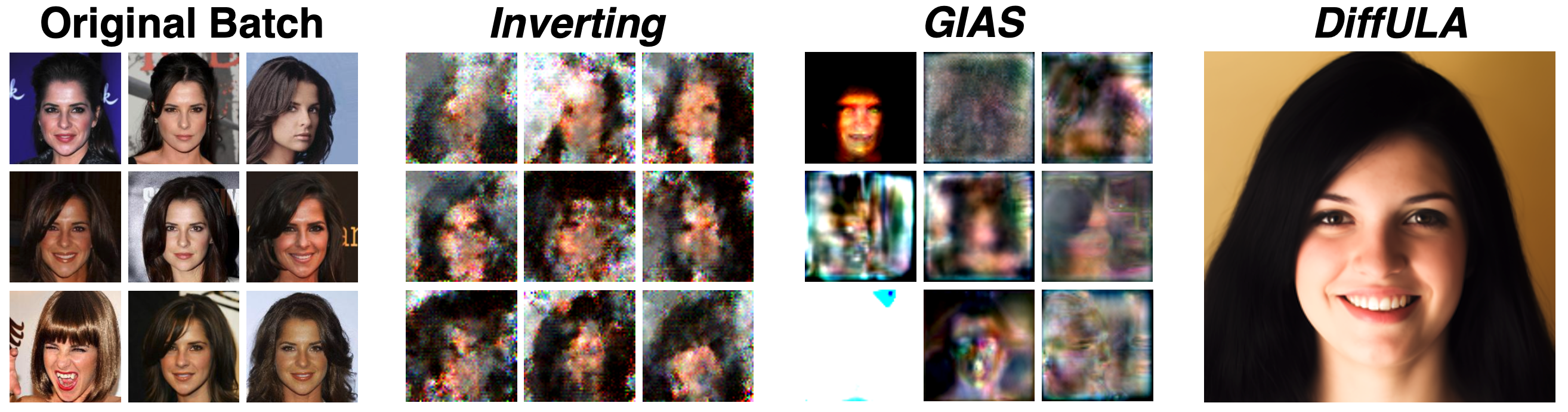}

    \caption{Visual comparison of reconstruction results from a batch of $30$ images.}
    \label{fig:visual_comp}

\end{figure}

\paragraph{Image Quality.}
Table~\ref{tab:res_quality} presents the results on reconstructed image quality with respect to the original data. For reference, we also report the average pairwise (intra-user) score measured across the original private image batch.
We observe that \methodName~achieves the best image quality scores compared to inverting and GIAS.
However, as the optimization problem in Eq.~\ref{eq:grad_match} is permutation-invariant, the reconstructed images by sample-level attacks are arbitrarily ordered, which creates ambiguity for comparing pair-wise image quality. To disambiguate, we apply the Jonker-Volgenant algorithm~\cite{crouse2016implementing} to solve for the optimal original-reconstructed image pairing that minimizes total MSE.
Although the measured MSE and PSNR of inverting and GIAS improved significantly after disambiguation and even surpassed the intra-user reference, this result does not agree with the visual comparison shown in Figure~\ref{fig:visual_comp} (see Appendix~\ref{sec:add_visual} for the full version).
Overall, LPIPS is the most interpretable metric that is better aligned with human perception.

\paragraph{Semantics.} Table~\ref{tab:res_seman} compares the performance in terms of facial semantics.
We also report the results on the original batch
as a reference for the experimental errors induced by the variation in the original images and the inaccuracies of the facial attribute predictors.
We notice that inverting and GIAS may generate extremely distorted images that fail to be detected by face detectors. To account for this effect, we employ a simple batch ensemble strategy to disregard images without a detected face within each batch for evaluation. When there is no single image with a detected face in the batch, we set the attack to random guessing. The results show that reconstructed images by inverting and GIAS have a very low sample-level face detection rate (3\% and 17\%).
Applying the ensemble strategy, 22\% of reconstructed batches by inverting and 96\% by GIAS have at least a single image with a detectable face. Although \methodName~is able to reliably generate realistic facial images (100\% detection rate), it shows no advantage over baselines in terms of recovering most facial attributes (except for gender). This indicates that the reconstructed images by inverting and GIAS might still contain useful cues for recovering private information despite their poor visual quality.

\paragraph{Computational Costs.} We measure the computational times for running the attack
on an Nvidia A100 GPU. Inverting gradients takes $23.2$ minutes to run $24{,}000$ optimization steps. GIAS takes $60.7$ minutes to run $1{,}000$ latent space search steps and $8{,}000$ parameter space search steps. In comparison, \methodName, which requires $4{,}000$ optimization steps, takes only $23.3$ minutes.
This computational advantage enables \methodName~to be applied to larger batch sizes that were previously hard to achieve with traditional methods (e.g., see Appendix~\ref{sec:large_batch} for results with batch size of 100).

\section{Conclusion}

This work investigates user-level gradient inversion as a new attack surface in distributed learning. We first examine existing attacks in this new scenario and reveal the inefficacy of the existing method on practical batch sizes. We then explore employing diffusion prior for efficient reconstruction of representative images from large batches of images under the user-level assumption.
Our study highlights the importance of designing pragmatic attacks that recover semantically meaningful private information under practical assumptions, as well as the need for better privacy metrics that are more aligned with human intuition,
which are both interesting avenues for future research.

\newpage

{
\small

\bibliography{reference}
\bibliographystyle{unsrt}
}

\newpage

\appendix

\section{Visualization}\label{sec:add_visual}

\begin{figure}[h]
    \centering
    \includegraphics[width=0.98\textwidth]{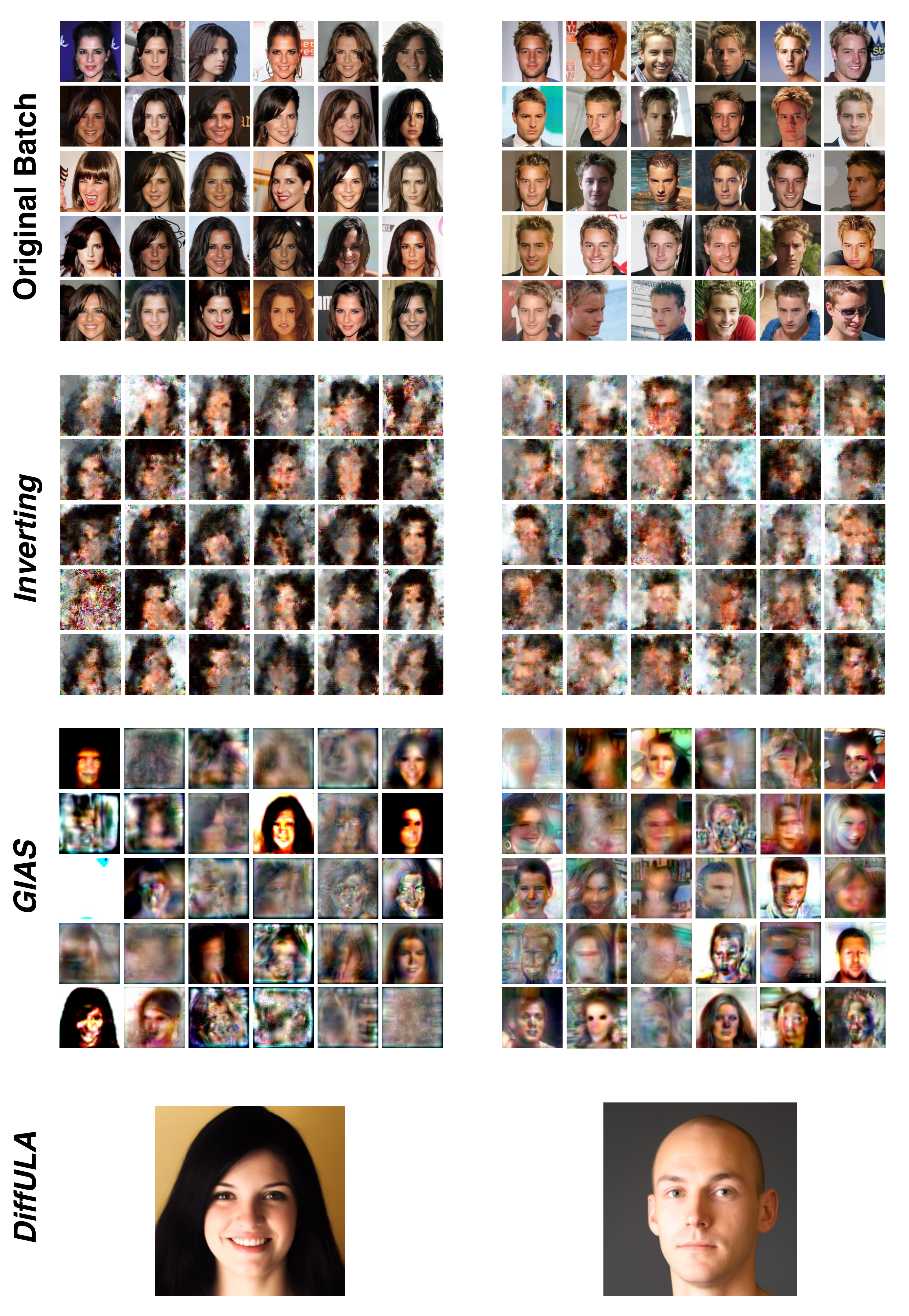}
    \vspace{-1mm}
    \caption{Visual comparison of reconstruction results (full batch).}
    \label{fig:visual_comp_full}
    \vspace{-2mm}
\end{figure}

\section{Results on Large Batch Size}\label{sec:large_batch}

\begin{figure}[h]
     \centering
     \begin{subfigure}[b]{0.8\textwidth}
         \centering
         \includegraphics[width=\textwidth]{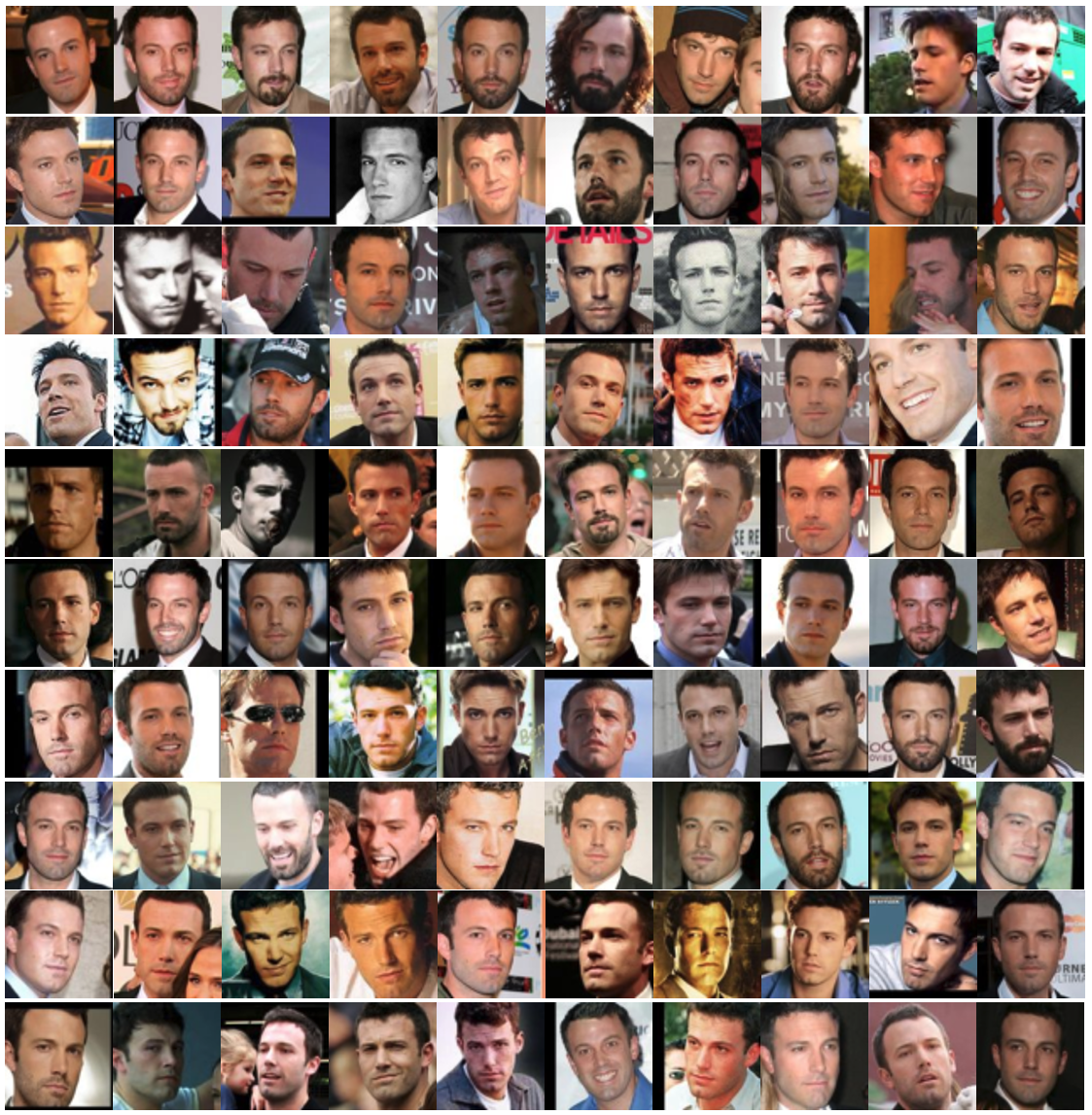}
         \caption{Original batch of 100 images}
         \label{fig:t_schedule}
     \end{subfigure}
     \begin{subfigure}[b]{0.94\textwidth}
         \centering
         \includegraphics[width=\textwidth]{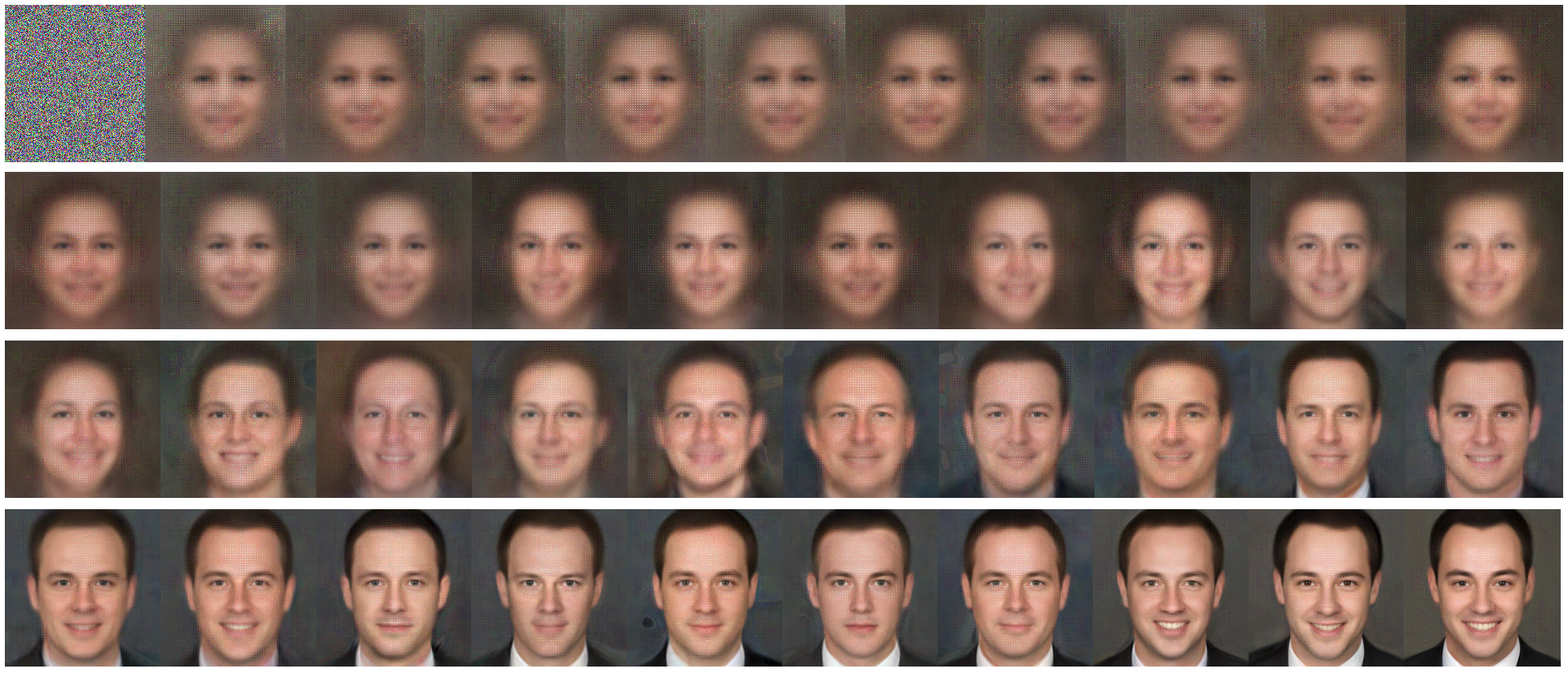}
         \caption{Reconstructed image of every 100 optimization steps}
         \label{fig:window}
     \end{subfigure}
     \begin{subfigure}[b]{0.24\textwidth}
         \centering
         \includegraphics[width=\textwidth]{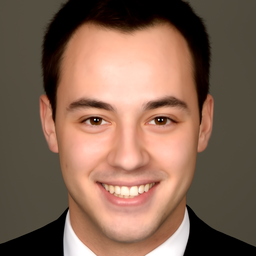}
         \caption{Denoised image}
         \label{fig:window}
     \end{subfigure}
     \vspace{-1mm}
     \caption{Example of reconstructing from a batch of 100 images. The measured pairwise facial similarity in the private batch ranges from 0.4371 to 0.9528. The average facial similarity of the reconstructed image to the original batch is 0.7080.}
     \label{fig:large_batch}
\end{figure}

To explore the feasibility of reconstructing from larger batch sizes using \methodName, we take the first 100 images (resized to $64\times64$ px) from the first 7 celebrities from the PubFig83 dataset~\cite{pinto2011scaling}. Figure~\ref{fig:large_batch} provides a visualization of the reconstruction and the numerical scores are summarized in Table~\ref{tab:large_batch}. We observe similar performance as with a batch size of 30.

\begin{table}[]
    \centering
    \caption{Numerical results with a batch size of 100.}
    \vspace{1mm}
    \resizebox{0.96\linewidth}{!}{
    \begin{tabular}{ccccccccc}
    \toprule
    & \textbf{MSE$\downarrow$} & \textbf{PSNR$\uparrow$} & \textbf{LPIPS$\downarrow$} & \textbf{\begin{tabular}[c]{@{}c@{}}Face\\ Detection Rate$\uparrow$\end{tabular}} & \textbf{\begin{tabular}[c]{@{}c@{}}Facial\\ Similarity$\uparrow$\end{tabular}} & \textbf{\begin{tabular}[c]{@{}c@{}}Gender\\ Accuracy$\uparrow$\end{tabular}} & \textbf{\begin{tabular}[c]{@{}c@{}}Race\\ Accuracy$\uparrow$\end{tabular}} &  \textbf{\begin{tabular}[c]{@{}c@{}}Age\\ Error$\downarrow$\end{tabular}} \\ \toprule
    \textit{\textbf{\methodName}}
    & 0.4028
    & 3.1416
    & 0.5101
    & 1.00                                                                            & 0.5859                                                                         & 0.57                                                                         & 0.86                                                                                                                                                 & 8.08                                                                     \\ \bottomrule
    \end{tabular}
    }
    \label{tab:large_batch}
\end{table}

\section{Implementation Details}

We employ a cosine-modulated linear annealing time schedule with added uniform noise similar to \cite{graikos2022diffusion}. As shown in Figure~\ref{fig:t_schedule}, $\tau$ is scheduled to start from $1,000$ and gradually decreases to $500$.
We balance the effects of recovering the user's private image and generating a realistic facial image by clipping the norm of the gradient of the gradient matching term. The clipping scale factor $\zeta$ is scheduled to follow a cosine ramp down from $1.5$ to $1.0$.
Additionally, as visualized in Figure~\ref{fig:window}, we apply a window function to focus on the deeper layers and gradually introduce the shallower layers to the computation of the gradient-matching term as the optimization progresses.

We implemented $\mathcal{T}$ as a series of random image transformations, including Gaussian noise injection, random color jitter, random perspective transformation, and Gaussian blur. Each transformation has a probability of $0.5$ to be applied for a given image. As the synthesized image from DDPM is of higher resolution, they are further resized before being fed into the learning model $f_{\rvw}$.

After optimization, we run some additional denoising steps on the reconstructed image using DDPM starting from $t^*=200$ to further improve image quality. The complete procedure of the proposed attack is presented in Algorithm~\ref{alg:user_level}.

\begin{figure}[h]
     \centering
     \begin{subfigure}[b]{0.42\textwidth}
         \centering
         \includegraphics[width=\textwidth]{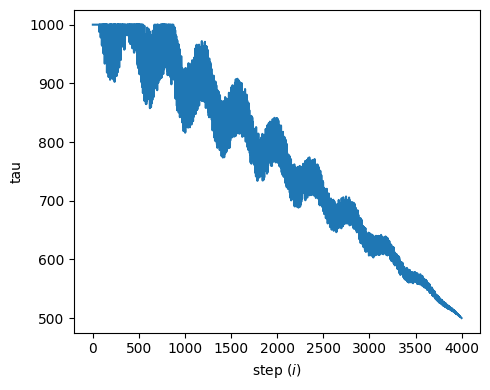}
         \caption{Time schedule}
         \label{fig:t_schedule}
     \end{subfigure}
     \hspace{5mm}
     \begin{subfigure}[b]{0.33\textwidth}
         \centering
         \includegraphics[width=\textwidth]{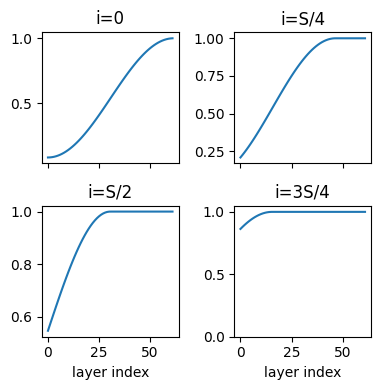}
         \caption{Window function}
         \label{fig:window}
     \end{subfigure}
     \caption{Time schedule and sliding window for optimization.}
\end{figure}

\begin{algorithm}[h]

\caption{User-level gradient inversion with diffusion prior}\label{alg:user_level}

\SetKwInOut{Input}{Input}
\SetKwInOut{Output}{Output}
\SetKwInOut{Parameter}{Parameter}
\SetKwComment{Comment}{/* }{ */}

\Input{observed gradient $\nabla \rvw$, model parameters ${\rvw}$, pretrained DDPM $\vepsilon_\vtheta$}

\Parameter{number of steps $S$, image transformation(s) $\mathcal{T}$, time schedule $\{\tau_i\}_{i=1}^S$, gradient clipping schedule $\{\zeta_i\}_{i=1}^S$, denoising timestep $t^*$, learning rate $\eta$}

Randomly initialize $\rvx_0 \sim \gN(\bm{0}, \mI)$\;
\For(\tcp*[h]{Optimization steps}){$i = 1, ..., S$}{
    Sample $\vepsilon \sim \gN(\bm{0}, \mI)$\;
    
    $\rvx_{\tau_i} \gets \sqrt{\bar{\alpha}_{\tau_i}}\rvx_0 + \sqrt{1-\bar{\alpha}_{\tau_i}}\vepsilon$\;

    $\rvg_\text{p} \gets \nabla_{\rvx_0} ||\vepsilon - \vepsilon_\vtheta(\rvx_{\tau_i}, \tau_i)||^2$ \Comment*[r]{Evaluate prior term}

    $\mathcal{A}(\rvx_0) \gets \{\mathcal{T}(\rvx_0) \}_{i=1}^B$\;

    $\rvg_\text{gm} \gets \nabla_{\rvx_0} \mathbf{d}\big(F(\mathcal{A}(\rvx_0)), {\nabla \rvw}\big)$ \Comment*[r]{Evaluate gradient matching term}

    $\rvg_\text{gm} \gets \rvg_\text{gm}/{\text{max}(1, \frac{||\rvg_\text{gm}||}{\zeta_{\tau_i}||\rvg_\text{p}||})}$\;

    $\rvx_0 \gets \rvx_0 - \eta \cdot (\rvg_\text{p} + \rvg_\text{gm}) $ \Comment*[r]{Update via gradient descent}
}

$\rvx_{t^*} \gets \rvx_0$\;

\For(\tcp*[h]{Denoising steps}){$t = t^*, ..., 1$}{
    \eIf{$t>1$}{
    Sample $\rvz \sim \gN(\bm{0}, \mI)$\;
    }{
    $\rvz \gets \bm{0}$\;
    }
    $\rvx_{t-1} \gets \frac{1}{\sqrt{\alpha_t}}\big( \rvx_t - \frac{1- \alpha_t}{\sqrt{1- \bar{\alpha}_t}}\vepsilon_\vtheta(\rvx_t, t) \big) + \sigma_t \rvz$\;
}

$\hat{\rvx} \gets \rvx_0$, $\rvh_c \gets \argmax p(\rvh_c|\rvx_0)$\;

\Return reconstructed image $\hat{\rvx}$, inferred $\rvh_c$
\end{algorithm}

\section{Discussion}

One inherent limitation of the proposed single-image-based, user-level gradient inversion is that it suffers from the ``regression toward the mean'' effect, that is, the best reconstruction of this strategy is to approximate the average of images in the batch. As such, its reconstruction performance highly depends on the image variance within the batch.

\begin{figure}[h]
    \centering
    \includegraphics[width=0.46\textwidth]{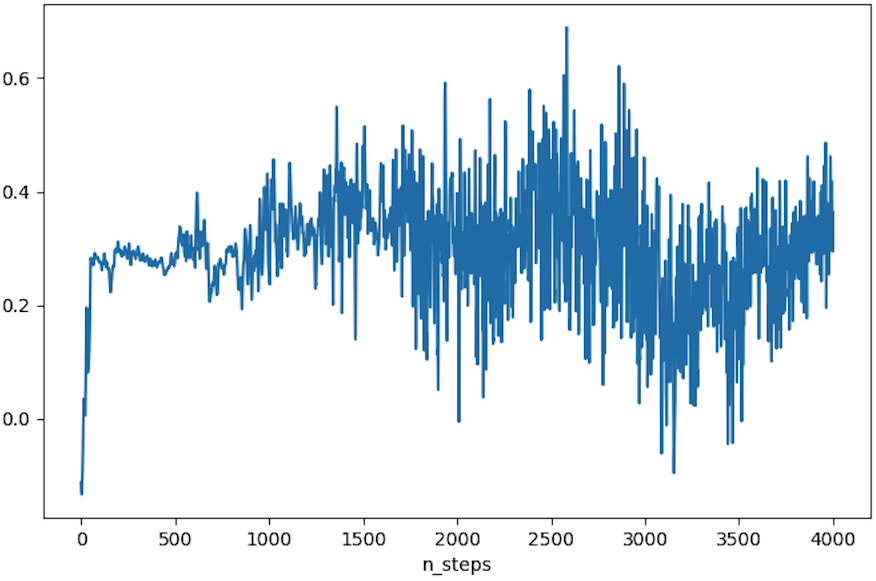}
    \vspace{-1mm}
    \caption{Facial similarity score per optimization step.}
    \label{fig:sim_per_step}
    \vspace{-2mm}
\end{figure}

Many previous works have reported observing varying reconstruction performance across different images, i.e., some images are innately more susceptible to gradient inversion than others (e.g., Figure 3 in~\cite{geiping2020inverting}). Currently, there is a lack of theoretical explanation for this phenomenon.
The diffusion approach may perform poorly if the target image is not well-represented in the prior distribution $p_\theta(\rvx_0)$, which is a common limitation shared with all reconstruction methods relying on a generative prior.
Additionally, in our experiments, we find that the utilization of diffusion prior may amplify this invariance due to the instability of reconstruction caused by the stochasticity of the diffusion process. For instance, as shown in Figure~\ref{fig:fail_cases}, we observe that in most of the failed cases, the reconstruction diverges when it approaches the end of the process. This is likely because the diffusion prior is prone to hallucinate when optimizing low $t$ values where it attempts to add details to the image. This can also be seen from the progression of the facial similarity score of the reconstructed image during the optimization process, as shown in Figure~\ref{fig:sim_per_step}. Adjusting the time schedule to only optimize for high $t$ values may improve the accuracy of the inferred private attributes, but would result in generating lower-quality images, i.e., there is an inherent trade-off between the reconstruction of high-level semantics and image details.

\begin{figure}[h]
    \centering
    \includegraphics[width=0.9\textwidth]{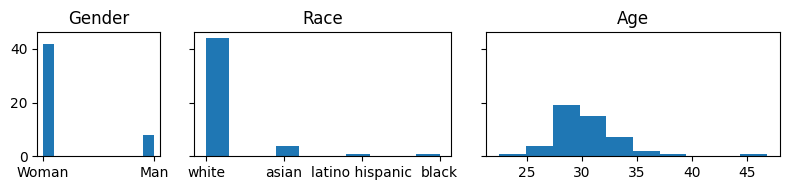}
    \vspace{-1mm}
    \caption{Demographics of the users involved in experiments.}
    \label{fig:user_demo}
    \vspace{-2mm}
\end{figure}

\begin{figure}[h]
    \centering
    \includegraphics[width=0.98\textwidth]{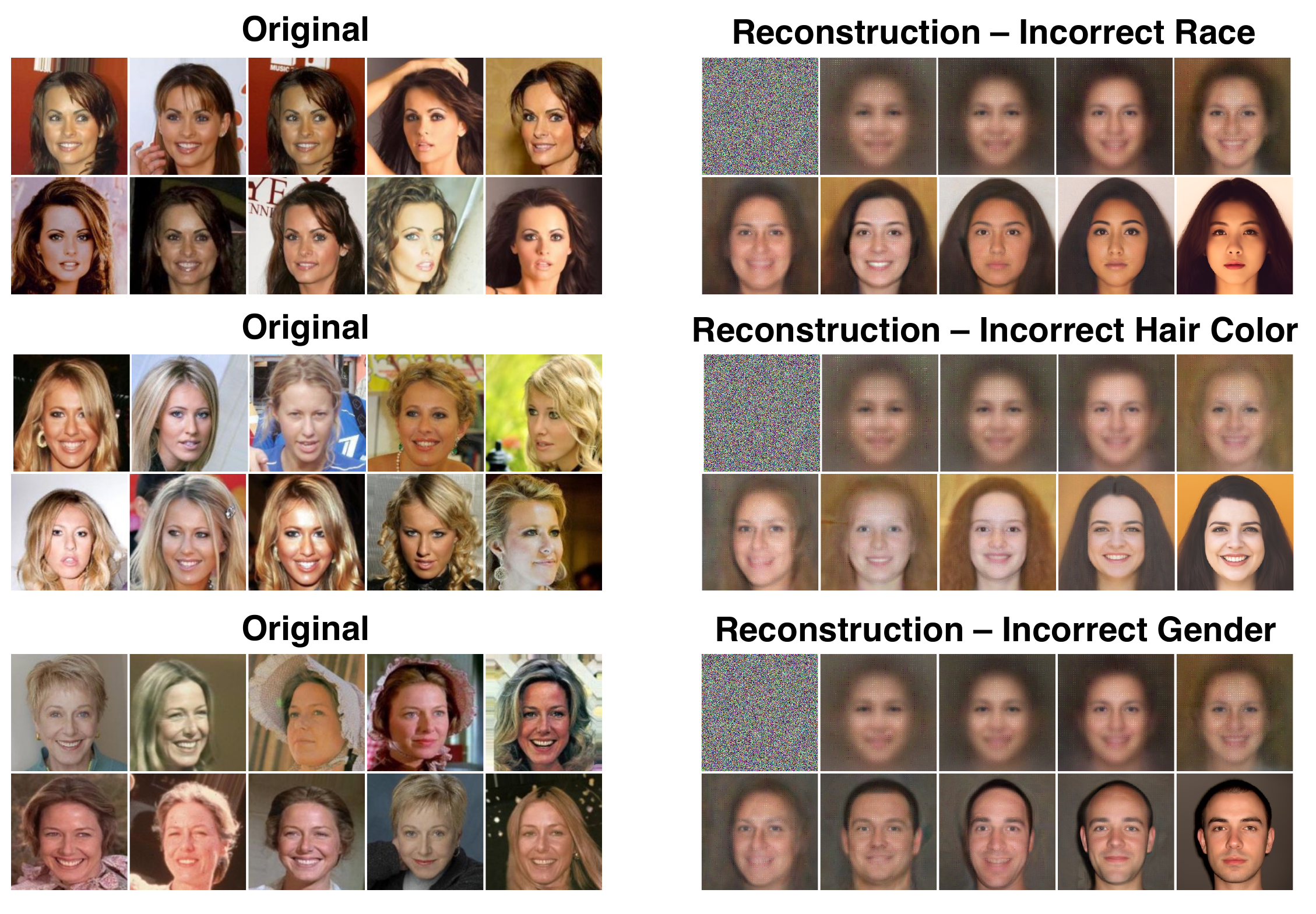}
    \vspace{-1mm}
    \caption{Reconstruction process of failed cases.}
    \label{fig:fail_cases}
    \vspace{-2mm}
\end{figure}

Regarding evaluation, we note that the facial semantics analysis relies on deep learning models that are trained on different data distributions which may introduce noise to the derived predictions. For instance, the face embedding model is trained on VGG Face data, while the DDPM is trained on the FFHQ dataset. In addition, the data selected for experiments are skewed, as shown in Figure~\ref{fig:user_demo}, which may cause the results to be biased toward certain demographics.

\section{Alternative Formulation}
There are several alternative ways to implement the user-level inversion as described in Eq.~\ref{eq:user_level_objective}.
For instance, apart from the single-image-based formulation explored in this paper, one may modify the sample-level approach to enforce the consistency of the latent encoding in the reconstructed images through the guidance of external classifiers.
Alternatively, an adversary with access to the private data distribution may learn a model to directly predict $\rvh_c$ and further leverage the recovered $\rvh_c$ as a condition for synthesizing private images, i.e., model $p(\rvh_c | {\nabla \rvw})$ and then sample from $p(\rvx|\rvh_c)\propto p_\vtheta(\rvx) p(\rvh_c|\rvx)$. Regarding the utilization of the diffusion prior, an alternative approach is to use guided sampling from diffusion with modified conditions~\cite{dhariwal2021diffusion,chung2023diffusion}.
We leave explorations in these spaces as future work.

\end{document}